# Deep learning based closed-loop optimization of geothermal reservoir production

Nanzhe Wang[a, b], Haibin Chang[c,*], Xiangzhao Kong[b], Martin O. Saar[b], and Dongxiao Zhang[d,*]

[a] BIC-ESAT, ERE, and SKLTCS, College of Engineering, Peking University, Beijing 100871, P. R. China
[b] Geothermal Energy and Geofluids Group, Department of Earth Sciences, ETH Zürich, 8092, Zürich, Switzerland
[c] School of Energy and Mining Engineering, China University of Mining and Technology (Beijing), Beijing 100083, P. R. China
[d] National Center for Applied Mathematics Shenzhen (NCAMS), Southern University of Science and Technology, Guangdong, P. R. China

*Corresponding authors: E-mail address: changhaibin@pku.edu.cn (Haibin Chang); zhangdx@sustech.edu.cn (Dongxiao Zhang)

**Abstract**

To maximize the economic benefits of geothermal energy production, it is essential to optimize geothermal reservoir management strategies, in which geologic uncertainty should be considered. In this work, we propose a closed-loop optimization framework, based on deep learning surrogates, for the well control optimization of geothermal reservoirs. In this framework, we construct a hybrid convolution-recurrent neural network surrogate, which combines the convolution neural network (CNN) and long short-term memory (LSTM) recurrent network. The convolution structure can extract spatial information of geologic parameter fields and the recurrent structure can approximate sequence-to-sequence mapping. The trained model can predict time-varying production responses (rate, temperature, etc.) for cases with different permeability fields and well control sequences. In the closed-loop optimization framework, production optimization based on the differential evolution (DE) algorithm, and data assimilation based on the iterative ensemble smoother (IES), are performed alternately to achieve real-time well control optimization and geologic parameter estimation as the production proceeds. In addition, the averaged objective function over the ensemble of geologic parameter estimations is adopted to consider geologic uncertainty in the optimization process. Several geothermal reservoir development cases are designed to test the performance of the proposed production optimization framework. The results show that the proposed framework can achieve efficient and effective real-time optimization and data assimilation in the geothermal reservoir production process.

# 1 Introduction

Geothermal energy is considered to be a new and highly promising source of clean energy, and has attracted increasing attention in recent years due to its abundance on our planet, high efficiency, and renewable nature (Bu et al., 2012; Randolph & Saar, 2011; Song et al., 2018; Zhang et al., 2019). The exploitation of geothermal energy can assist to alleviate the energy shortage and the environmental pressure caused by using fossil fuel energy. The development of geothermal energy usually comprises injecting working fluids (e.g., water, $CO_2$, etc.) into the geothermal reservoirs and pumping the fluids out to extract heat from the reservoirs, which are further converted to electrical energy on the Earth's surface (Bu et al., 2012). The production process requires a high cost, including well drilling, fluid injection and pumping, etc., and thus it is essential to optimize geothermal reservoir management to obtain maximum economic benefits. Well control operation optimization constitutes one of the important reservoir management tasks in the geothermal energy development process.

Production optimization of geothermal energy is usually carried out based on numerical simulators. Various numerical methods, including the finite difference method (FDM), finite element method (FEM), finite volume method (FVM), etc., have been widely used to model the interaction between subsurface fluids and rocks in geothermal reservoirs (Pandey et al., 2018), which also serve as effective tools to understand the heat extraction process. However, heat extraction from geothermal reservoirs is often coupled with multiple processes (Pandey et al., 2018; Saar et al., 2015), such as hydro, thermo, mechanical, and chemical processes, which would be computationally demanding to solve numerically. Moreover, the numerical solvers need to be run a large number of times to evaluate the objective function during the optimization process of reservoir operations, which would further increase computational burden (Biagi et al., 2015; Blank et al., 2021). Instead of using numerical solvers directly, constructing surrogate models of numerical simulators constitutes a new way to accelerate the optimization process (Forrester & Keane, 2009; Queipo et al., 2005).

Surrogate models can approximate the relationship between the well operation parameters and production data to replace the numerical solving processes of partial differential equations, which would be much more efficient to evaluate. Various kinds of surrogate models have been used in geofluids flow and geothermal energy problems, including polynomial expansion, radial basis function, support vector machine, multivariate adaptive regression spline (MARS), etc. (Asai et al., 2018; Chang et al., 2017; Chen et al., 2015; Christelis & Mantoglou, 2016; He et al., 2019). Chen et al. (2015) employed the MARS technique to construct a statistical surrogate model for high-fidelity geothermal models, which was utilized for parameter sensitivity quantification, well placement, and control optimization of geothermal reservoirs. Asai et al. (2018) analyzed the effects of parameters, including well spacing, fracture spacing, well inclination angle, injection temperature, and injection rate towards the performance of an enhanced geothermal system (EGS), in which a second-order polynomial surrogate was constructed for the sensitivity study. Babaei et al. (2022) constructed a surrogate response surface model for efficient doublet well spacing optimization in geothermal systems. Different from traditional methods, deep learning approaches can provide an alternative way for surrogate modeling of geothermal systems due to their more powerful function approximation capability.

Deep learning has received increasing attention because of advancements in computer hardware, such as graphics processing units (GPUs), which provide powerful computing resources and speed up calculations substantially. Due to its advantages, deep learning algorithms have also been used in modeling subsurface flow and energy exploitation problems. For example, deep learning models can be trained to predict pressure fields, temperature fields, stress fields, etc., in geothermal reservoirs with available training data (Aydin et al., 2020; Wang et al., 2020). Bassam et al. (2015) proposed a new approach of artificial neural network (ANN) techniques for the determination of pressure drops between geothermal wells of inclined and vertical geometries. Tut Haklidir and Haklidir (2019) developed a deep neural network (DNN) model to predict geothermal fluid temperatures based on hydro-geochemistry data from geothermal reservoirs. Deep learning models can also be developed as surrogates for numerical simulators, which can be used to efficiently perform uncertainty quantification (Mo et al., 2019; Wang et al., 2021b), inverse modeling (Tang et al., 2020), and optimization design (Akın et al., 2010; Kim et al., 2020).

In this work, a deep learning surrogate-based closed-loop optimization framework is proposed for geothermal reservoir production, in which the production optimization and data assimilation procedures are implemented alternately to achieve real-time optimization of the geothermal reservoir. Because the subsurface geologic properties of the reservoirs (e.g., permeability, porosity, etc.) are usually uncertain, data assimilation can be first implemented to estimate the reservoir properties. In this work, the iterative ensemble smoother (IES) is adopted to update the geologic parameters. Production optimization can then be performed based on initial estimations of the geologic parameters to find the optimal well control operations for the next production period. The differential evolution (DE) algorithm is adopted in the optimization part. Considering that the estimations of reservoir properties may also contain some uncertainties, geologic uncertainty should also be considered in the optimization process by evaluating the average objective function for the group of possible reservoir parameters. The optimized well controls can then be operated on the production fields, which would yield new production data. Data assimilation can then be performed again with the new available data to provide a more accurate estimation of reservoir parameters and further reduce geologic uncertainty. Optimization can then be implemented with the newly updated reservoir parameters for the subsequent production period. In this way, a closed loop can be formed to integrate the data assimilation and production optimization process simultaneously, and lead to optimal exploitation of geothermal energy. Furthermore, in order to improve the efficiency of optimization, a deep learning based surrogate is constructed, which combines the convolution neural network and long short-term memory (LSTM) recurrent network. The convolution layers in the model can effectively extract spatial information of geologic properties and transform them into latent variables. The LSTM part can learn sequence-to-sequence mapping from time-dependent well control sequences and time series of production data. The deep learning model can achieve satisfactory approximation accuracy after being trained with simulation datasets. Several geothermal reservoir development cases are designed to test the performance of the proposed framework. The effectiveness of deep learning surrogate-DE-based well control optimization is firstly validated with a case with a known permeability field, in which geologic uncertainty is not considered and the data assimilation part is not involved. Closed-loop optimization is then implemented on a geothermal reservoir case with an unknown permeability field, in which estimation of the permeability field and well control optimization can be realized simultaneously as production proceeds. The results demonstrate that the closed-loop optimization framework can provide a beneficial well control solution. In addition, more accurate estimations of the permeability field with reduced uncertainty can be obtained as more production data are assimilated.

The remainder of this paper is organized as follows. In section 2, the structure of the hybrid convolution-recurrent neural network, the methodology of IES and DE, and the framework of closed-loop optimization are introduced. In section 3, several two-dimensional geothermal energy production cases are considered to test the performance of the proposed optimization framework. Finally, discussions and conclusions are given in section 4.

## 2 Methodology

In this section, the deep learning surrogate model, the iterative ensemble smoother algorithm (IES), the differential evolution algorithm (DE), and the closed-loop optimization workflow are introduced.

### 2.1 Hybrid convolution-recurrent neural network surrogate

In this work, a deep learning based surrogate model is constructed for the geothermal production problem to approximate the relationship between permeability fields, well control sequences, and production data. Considering that both time series data and image data are involved, a hybrid convolution-recurrent neural network is constructed with a convolutional architecture to learn low dimensional representations of images and a recurrent architecture to learn sequence-to-sequence mapping, which has been demonstrated to be effective for spatial-temporal regression (Ma et al., 2021). The structure of the constructed surrogate model is presented in **Figure 1**. The images of permeability fields are encoded by convolutional layers into low dimensional latent variables, which are then decoded to reconstruct the images of permeability fields through deconvolution layers. The reconstruction loss can be calculated as follows:

$$L_{recon}(\theta_c) = \frac{1}{N_k} \sum_{i=1}^{N_k} \left\| \ln K_i^{recon} - \ln K_i^{input} \right\|_2^2, \tag{1}$$

where $\theta_c$ denotes the parameters (weights and bias) of the convolutional encoder-decoder network; and $N_k$ denotes the number of permeability fields utilized to train the model. Minimizing reconstruction loss can regulate the model to compress the spatial permeability fields more effectively, and thus the extracted latent variables can represent the inputted images sufficiently and can be used as more valuable information when constructing the sequence-to-sequence mapping.

The compressed latent variables are replicated to the length of well control sequences, and the replicated latent variable sequences can then be stacked together with the well control sequences of different wells. Therefore, the new combined sequence has now incorporated both the well control information and the spatial permeability field information. The combined sequence is further fed into the recurrent network to predict production sequence data, such as production rate, well temperature, etc. In this work, a long short-term memory (LSTM) recurrent network is adopted to learn the sequence-to-sequence mapping, which is an established and widely used recurrent network variant (Hochreiter & Schmidhuber, 1997; Jia et al., 2015; Palangi et al., 2016; Zhang et al., 2017). Compared to the conventional recurrent network, LSTM can effectively avoid the gradient vanishing and explosion problems, especially for long sequences, and learn long-term dependence information due to its specially designed gate units. A more detailed introduction of LSTM can be found in Hochreiter and Schmidhuber (1997). The sequence mismatch loss can be calculated as follows:

$$L_{seque}(\theta_r) = \frac{1}{N_k N_t} \sum_{i=1}^{N_k} \sum_{j=1}^{N_t} \left\| P_{i,j}^{pred} - P_{i,j}^{true} \right\|_2^2, \quad (2)$$

where $\theta_r$ denotes the parameters of the LSTM network; $N_t$ denotes the total length of the sequences; and $P$ denotes the production data, e.g., production rates, production temperature, BHP, etc. Therefore, the total loss function of the hybrid deep learning model can be expressed as:

$$L(\theta_c, \theta_r) = L_{recon}(\theta_c) + L_{seque}(\theta_r). \quad (3)$$

It is also worth noting that the convolution part and the recurrent part of the deep learning model are trained simultaneously by minimizing the total loss function Eq. (3).

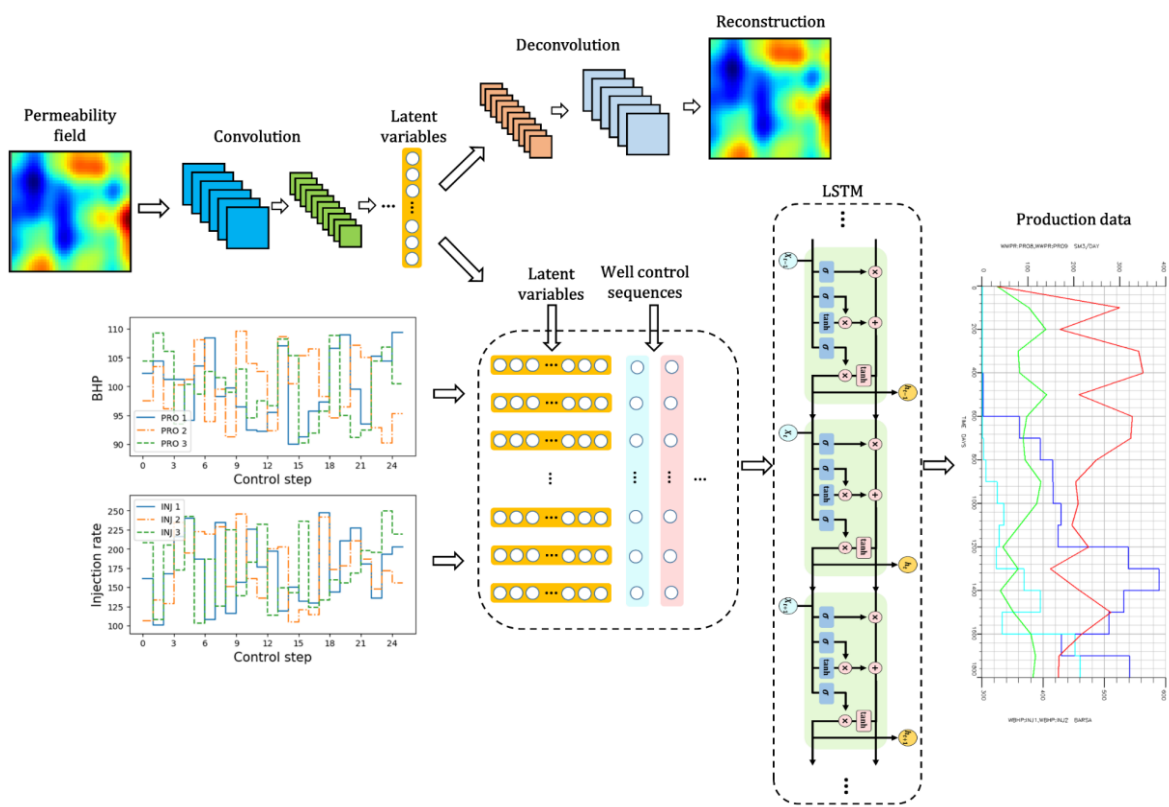

**Figure 1.** Structure of the hybrid convolution-recurrent neural network surrogate.

## 2.2 Iterative ensemble smoother-based data assimilation

In the optimization process, the newly obtained production data can be employed for real-time data assimilation to estimate the reservoir properties and reduce geologic uncertainty. In this work, the iterative ensemble smoother (IES) algorithm is adopted for data assimilation, which is an ensemble-based inverse modeling method, and has been used for history matching in oil reservoir simulation (Chang et al., 2017; Chen & Oliver, 2013; Wang et al., 2021a). The static reservoir parameters (e.g., permeability) are updated iteratively to reduce the mismatch between the collected observation data and predictions from geothermal reservoir models. In the IES algorithm, a group of model parameter realizations (ensemble) are updated simultaneously, and the update scheme is formulated as follows:

$$\begin{aligned}\mathbf{m}_{l+1,j} = \mathbf{m}_{l,j} &- \frac{1}{1+\lambda_l}\left[C_{M_l} - C_{M_lD_l}\left((1+\lambda_l)C_D + C_{D_lD_l}\right)^{-1}C_{D_lM_l}\right]C_M^{-1}\left(\mathbf{m}_{l,j} - \mathbf{m}_j^{pr}\right) \\ &- C_{M_lD_l}\left((1+\lambda_l)C_D + C_{D_lD_l}\right)^{-1}\left(g(\mathbf{m}_{l,j}) - \mathbf{d}_j^{obs}\right), \quad j = 1,\ldots,N_e\end{aligned} \quad (4)$$

where $\mathbf{m}$ denotes the model parameters, which refers to permeability in this work; $l$ denotes the iteration number index; $j$ denotes the index of realizations in the ensemble; $N_e$ denotes the total number of realizations in the ensemble; $\lambda_l$ denotes a multiplier to adjust the update step-size; $C_{M_l}$ denotes the covariance matrix of model parameters at the $l$th iteration; $C_D$ denotes the covariance matrix of observation error, i.e., the production data measurement error in this work; $C_{M_lD_l}$ denotes the cross covariance matrix between the updated model parameters and observation data predictions at the $l$th iteration; $C_{D_lD_l}$ denotes the covariance matrix of observation data predictions; $\mathbf{m}^{pr}$ denotes the prior estimation of model parameters; $\mathbf{d}^{obs}$ denotes the true observation data; and $g(\mathbf{m})$ denotes the forward calculation model with model parameter $\mathbf{m}$, which usually refers to reservoir numerical simulators. Calculation of the gradients of objective functions when nonlinear numerical solvers are involved in the objective functions is usually challenging, which may limit the application of the gradient-based data assimilation method. In this ensemble-based method, however, calculation of the sensitivity matrix or gradients can be avoided and replaced with statistical information of the ensemble of realizations to overcome this problem. Since the model parameters need to be updated iteratively, a large number of numerical simulations are usually required, which may impose a large computational burden. Therefore, in this work, the geothermal numerical simulator is replaced by the former constructed deep learning based surrogate model to accelerate the data assimilation process.

**2.2 Differential evolution-based optimization**

In this work, the differential evolution (DE) algorithm is used for well control optimization, which is a heuristic stochastic optimization method and does not require the gradient information of objective functions (Storn & Price, 1997). Originated from the ideas of the genetic algorithm (GA), DE also simulates the biological evolutionary process in nature, and selects individuals that adapt to the environments iteratively to achieve optimization. DE has three main operations, including mutation, crossover, and selection. In the mutation process of DE, the vector difference of two randomly selected individuals from the population are scaled and summed with the third individual to produce the new mutation individual, which constitutes the main difference between DE and GA. The procedures of the DE optimization process are as follows:

1. Population initialization: Assume that there are $N_{ind}$ individuals in the population, with each being an $n$ dimensional vector. The initial population matrix can be randomly generated in the solution space with:

$$x_{i,j}(0) = x_j^l + rand(0,1)*(x_j^u - x_j^l), \quad (i=1,2,\cdots,N_{ind}, j=1,2,\cdots,n), \tag{5}$$

where $i$ and $j$ denote the index of individuals and vector dimensions, respectively; $x_j^l$ and $x_j^u$ denote the lower and upper bounds of the $j$th dimension, respectively; and $rand(0,1)$ denotes the random number generated from a uniform distribution $U[0,1]$.

2. Mutation: Consider the $g$th generation, and three different individuals are randomly sampled from the population ($x_{r1}(g)$, $x_{r2}(g)$, $x_{r3}(g)$). The newly generated mutant for the $i$th

target individual can be expressed as:

$$v_i(g) = x_{r1}(g) + F*(x_{r2}(g) - x_{r3}(g)), \tag{6}$$

where $F$ denotes the scaling factor, which controls the importance of difference vector $x_{r2}(g) - x_{r3}(g)$ in the differential mutation process. Here, the randomly sampled individual vector $x_{r1}(g)$ is called the base vector. According to the different sampled base vector and number of difference vectors, the mutation strategy can be represented as DE/rand/1, DE/best/1, DE/target/1, etc. In these expressions: rand represents the randomly sampled base vector; best represents that the base vector is the best individual of the current population; and target means that the base vector is the same as the target vector (Price, 2013). The number 1 represents the number of difference vectors utilized to calculate the mutant.

3. Crossover: In this operation, each individual crosses-over with its generated mutation individual, with each dimension in the trial vector $u$ choosing the original ones or the mutants according to a certain probability, which can be expressed as:

$$u_{i,j}(g) = \begin{cases} v_{i,j}(g), & \text{if } rand(0,1) \leq Cr \text{ or } j = j_{rand} \\ x_{i,j}(g), & \text{otherwise} \end{cases}, \tag{7}$$

where $Cr$ denotes the crossover control variable, which controls the probability that the trial vector chooses the parameters from the mutant vector; $j_{rand}$ denotes a random component of the mutant vector, which can ensure that the generated trial vector has at least one dimension inherited from the mutant vector.

4. Selection: According to the value of fitness function, the next generation $x_i(g+1)$ can be selected from the target individuals $x_i(g)$ and the trial individuals $u_i(g)$, which can be given as follows:

$$x_i(g+1) = \begin{cases} u_i(g), & \text{if } f(u_i(g)) \leq f(x_i(g)) \\ x_i(g), & \text{otherwise} \end{cases}, \tag{8}$$

where $f$ denotes the fitness function.

In the DE-based optimization process, the population evolves from generation to generation through the above introduced mutation, crossover, and selection operations, which would repeat until the algorithm reaches the maximum iteration number $G_{\max}$, or the optimal solution reaches a predetermined error threshold.

**2.4 Closed-loop optimization workflow**

In order to achieve efficient real-time optimization and data assimilation in the geothermal reservoir production process, a closed-loop optimization framework with deep learning surrogate-accelerated is constructed. The optimization, data assimilation, and geothermal reservoir are coupled together to form a closed-loop, through which the data and information sharing can be achieved to facilitate the optimization process. The closed-loop workflow is presented in **Figure 2**. The main procedures of the production optimization framework are as follows:

1. History production data collection: The historical production data of geothermal reservoirs, such as production rate, bottom hole pressure, well temperature, etc., can be collected firstly, and prepared for data assimilation.

2. IES-based data assimilation: The production data can then be employed to estimate the subsurface formation properties (e.g., permeability, porosity, etc.) with the IES algorithm. Because the IES algorithm is an ensemble-based method, a group of parameter realizations that can represent the uncertainty of the geologic properties would be updated simultaneously. The updated realizations with reduced uncertainty would be close to the true values.

3. DE-based production optimization: With the estimation of formation parameters, well control optimization for the next production period can then be implemented with the DE algorithm. In order to take geologic uncertainty into consideration, an average objective function of the group of parameter estimations can be evaluated and optimized.

4. Form a closed-loop: After the optimization process, the optimized well control operations can be implemented to the subsequent production period of the geothermal reservoir. The newly collected production data of the next period can be utilized for data assimilation to further calibrate the reservoir parameters and reduce geologic uncertainty. With the updated reservoir parameters, well control optimization of the next production period can be implemented. In this way, a closed-loop optimization workflow is established, which can effectively integrate the data assimilation and production optimization process simultaneously.

In the closed-loop optimization process, the optimal well operations of different control steps can be determined, and characterization of geothermal reservoir properties becomes increasingly clear as the geothermal reservoir production progresses. It is also worth mentioning that the constructed deep learning surrogate model is utilized for forward calculation to replace the numerical simulators in both the data assimilation and optimization procedures, which can significantly improve the efficiency of the algorithm.

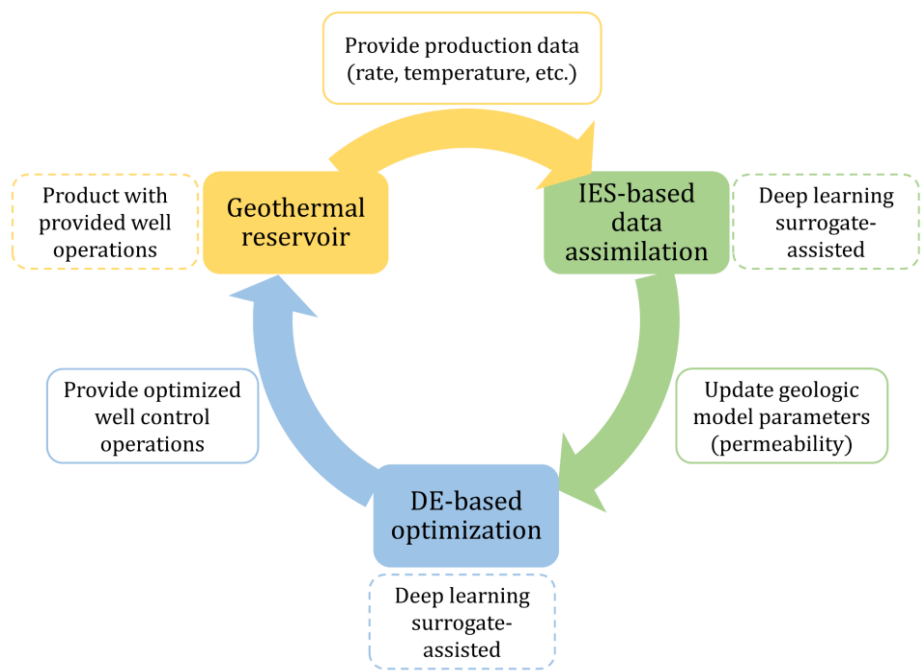

**Figure 2.** Workflow of the closed-loop optimization framework.

## 3 Case Studies

In this section, the performance of the proposed closed-loop optimization framework with deep learning surrogate is tested with several geothermal reservoir development cases.

## 3.1 DE-based production optimization with deep learning surrogate

In this subsection, production optimization by selecting the most rewarding well controls is performed for a geothermal reservoir, in which the geologic properties (permeability field in this work) is assumed to be known. Therefore, the data assimilation part can be omitted, and the DE-based optimization part can be implemented directly with the deep learning surrogate. This case is used to verify the effectiveness of DE and deep learning surrogate-based production optimization.

In this case, a 2D geothermal reservoir with square area $1220 \text{ m} \times 1220 \text{ m}$ is considered. The thickness of the reservoir is 30 m, and the initial reservoir temperature and pressure are 200 °C and 380 bar, respectively. The four boundaries are no-flow and insulated boundaries. In order to exploit the geothermal energy, cold water is injected into the reservoir and hot water is pumped out for energy conversion. In this case, four injection wells are drilled to inject cold water (20 °C), and five production wells are drilled to pump hot water out of the reservoir. The locations of the wells are presented in **Figure 3** (a). The injection wells are controlled by water injection rate, which ranges from 1100 $\text{m}^3/\text{day}$ to 1300 $\text{m}^3/\text{day}$. The production wells are controlled by bottom hole pressure (BHP), the operation range of which is between 230 bar and 250 bar. The thermal conductivity of water and rock are assumed to be 0.6 and 2 $\text{W}/\text{m}/\text{K}$, respectively. The porosity of the reservoir is set to be 0.1. The permeability field used in this case is generated with Karhunen-Loeve expansion (KLE) (Zhang & Lu, 2004), which is presented in **Figure 3** (b) and follows these statistics:

$$\langle \ln K \rangle = 3.6, \tag{9}$$

$$\sigma_{\ln K} = 1.0, \tag{10}$$

where $\langle \ln K \rangle$ denotes the mean, and $\sigma_{\ln K}$ denotes the standard deviation, of the log-transformed permeability field, $\ln K$.

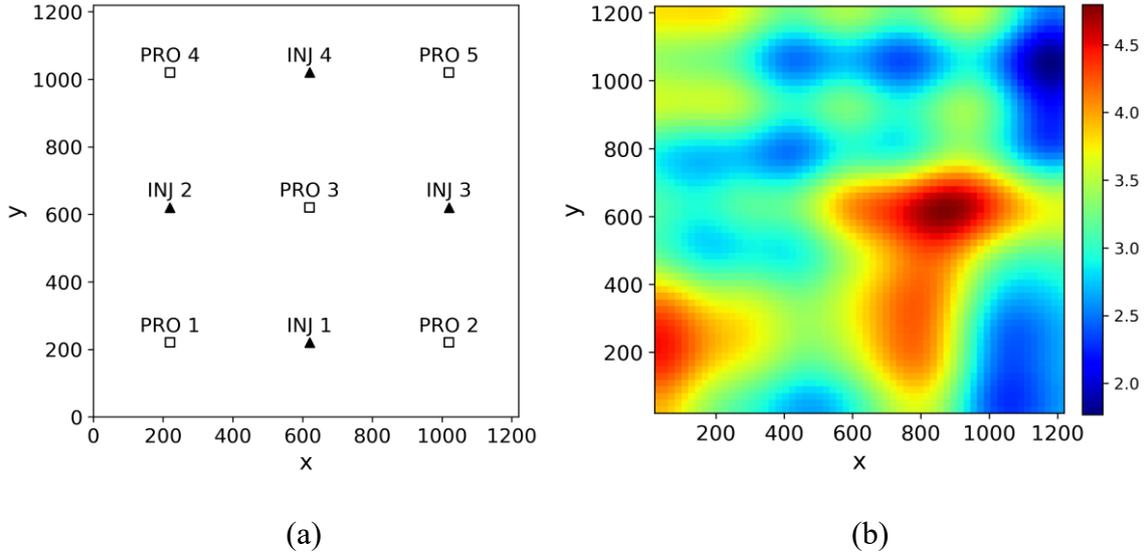

**Figure 3.** (a) Well locations in the geothermal reservoir. The filled triangles denote the injection wells, and the empty squares denote the production wells. (b) The permeability field of the geothermal reservoir.

The geothermal reservoir model can be solved with numerical simulator MATLAB Reservoir Simulation Toolbox (MRST) (Lie, 2019), in which the reservoir is divided into $61 \times 61$ grid blocks with each of size 20 m $\times$ 20 m. The total simulation time span is 10.27 years and the well control is adjusted every 150 days, and thus there would be 25 control steps in the entire simulated production period. The MRST simulator can be utilized to provide training datasets for deep learning surrogate construction.

### 3.1.1 Constructing convolution-recurrent neural network surrogate

The deep learning surrogate is firstly constructed in this subsection. Considering that the geologic model of the reservoir is assumed to be known in this case, only the well operation schedules serve as model inputs of the surrogate model. To provide the training dataset of the surrogate model, 500 groups of well control sequences for the producers and injectors are randomly generated, and the production data sequences are solved with the numerical simulator. The deep learning surrogate model is then trained with these paired sequence-to-sequence datasets. In addition, the same permeability fields are inputted into the model with different well control sequences. The deep learning surrogate model is trained for 3,000 epochs with a learning rate of 0.001, which takes approximately 600 s on the NVIDIA TITAN RTX. In order to test the accuracy of the trained surrogate, 100 more groups of well control sequences are generated and inputted into the numerical simulator to solve the corresponding production data, which can be employed as testing datasets.

A group of well control sequences is randomly sampled from the testing datasets, as shown in **Figure 4**. For these well control operations, the corresponding production data predicted with the deep learning surrogate, as well as numerical references, including production rate, temperature and BHP, are shown in **Figure 5**, **Figure 6**, and **Figure 7**, respectively. It can be seen that the predictions from the surrogate model can match the numerical solutions well, which demonstrates the satisfactory accuracy of the constructed surrogate model. To evaluate the accuracy of the surrogate statistically, the scatter plots of production data from predictions and references at two time nodes are presented in **Figure 8**. It can be seen that the scatters converge near straight lines with a 45° angle, which illustrates the accuracy of the surrogate for different well control sequences.

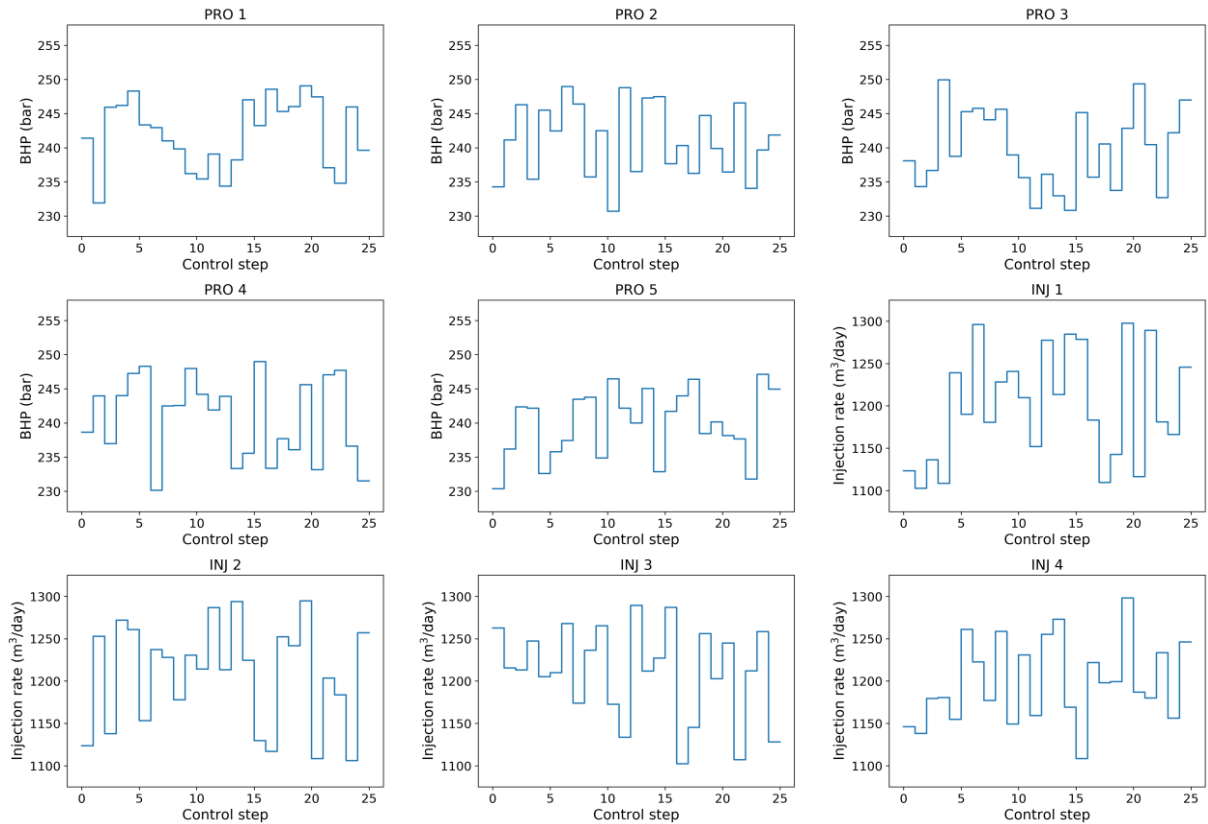

**Figure 4.** Sampled well control sequences from the testing datasets.

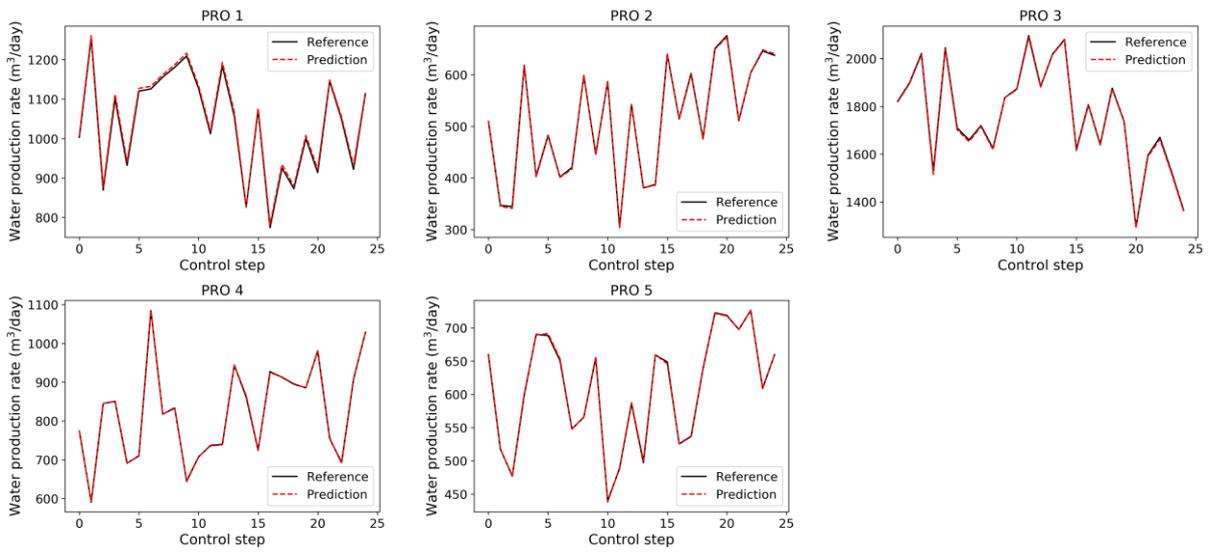

**Figure 5.** Comparisons between predictions from the deep learning surrogate and numerical solutions of water production rate.

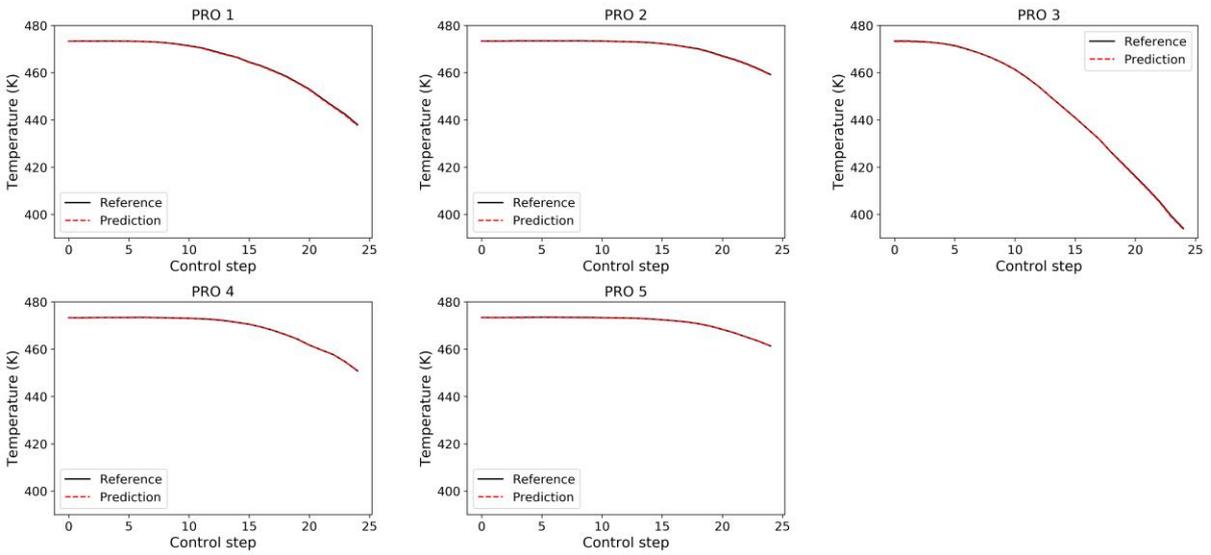

**Figure 6.** Comparisons between predictions from the deep learning surrogate and numerical solutions of production temperature.

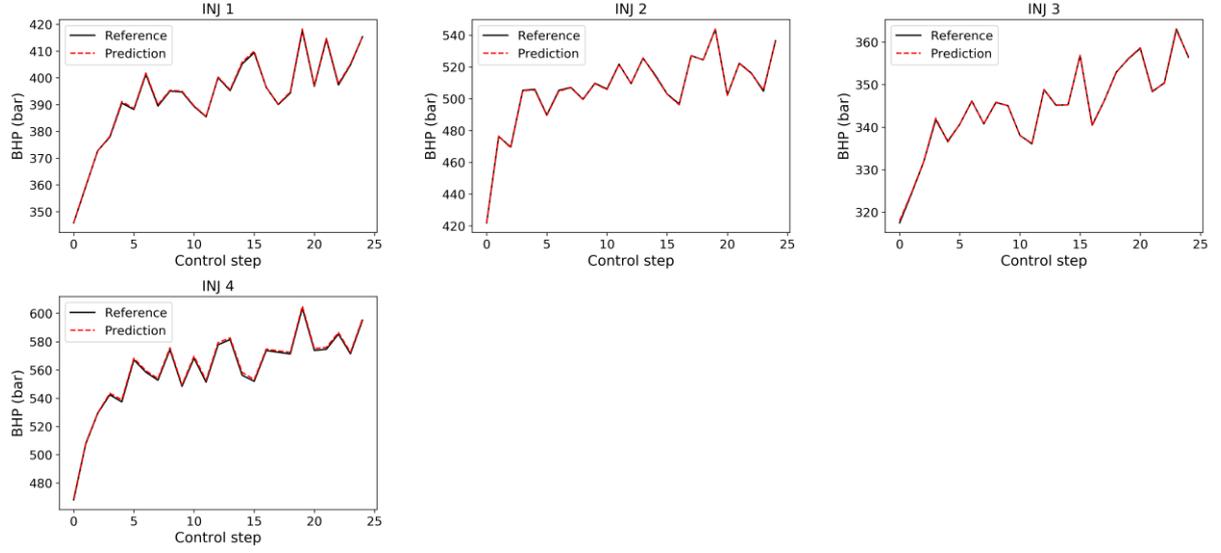

**Figure 7.** Comparisons between predictions from the deep learning surrogate and numerical solutions of BHP.

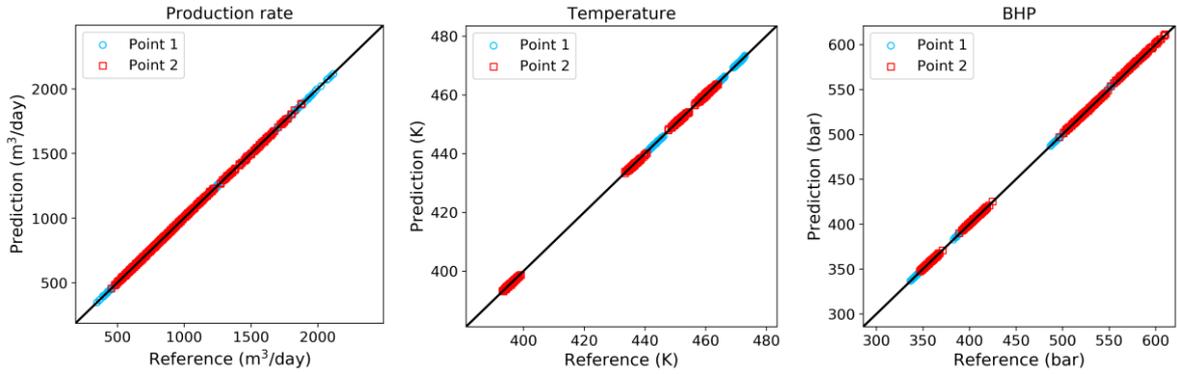

**Figure 8.** Scatter plots of production responses from predictions and references.

### 3.1.2 DE-based production optimization

The trained deep learning surrogate can then be used for well control optimization for geothermal reservoir production to maximize economic benefits. In this case, the permeability field is assumed to be known, and consequently only the well control sequences need to be adjusted in the optimization process.

In order to maximize the economic benefits of geothermal energy production, the net present value (NPV) is utilized as the objective function of optimization, which can be expressed as follows:

$$O(\mathbf{x}) = \sum_{t=1}^{T} \frac{\left( C_e E_{prod}^t - C_{wp} Q_{wp}^t T^t - C_{wi} Q_{wi}^t T^t \right)}{(1+d)^t}, \tag{11}$$

where $\mathbf{x}$ denotes the parameters to be optimized, i.e., well control sequences in this case; $T$ denotes the total number of time steps in the whole production period; $d$ denotes the annual discount rate; $C_e$ denotes the price of thermal energy per megawatt-hour (MWh); $C_{wp}$ and $C_{wi}$ denote the costs of water production and injection per cubic meter, respectively; $Q_{wp}^t$ and $Q_{wi}^t$ denote the water production and injection volume at time step $t$, respectively; $T^t$ denotes the time step size of the $t$th time step; and $E_{prod}^t$ denotes the thermal energy production at time step $t$, which can be calculated as follows:

$$E_{prod}^t = \rho_f C_f \left( \sum_{j=1}^{N_{pro}} Q_{wp_j}^t T^t - \sum_{k=1}^{N_{inj}} Q_{wi_k}^t T^t \right), \tag{12}$$

where $\rho_f$ and $C_f$ denote the density and specific heat capacity of working fluids (water in this case), respectively; and $N_{pro}$ and $N_{inj}$ denote the number of production and injection wells in the geothermal reservoir, respectively.

The DE algorithm is adopted to perform the optimization task. In addition to maximizing the NPV, other constraints can also be considered. For instance, in this case, we assume that the temperature at production wells should be higher than a critical value $T_c$ to make the motors convert heat energy into electricity effectively, which can be expressed as:

$$T_{pro,i} > T_c, \quad i = 1, 2, \cdots, N_{pro}. \tag{13}$$

The critical temperature $T_c$ is set to be 130 °C in this case.

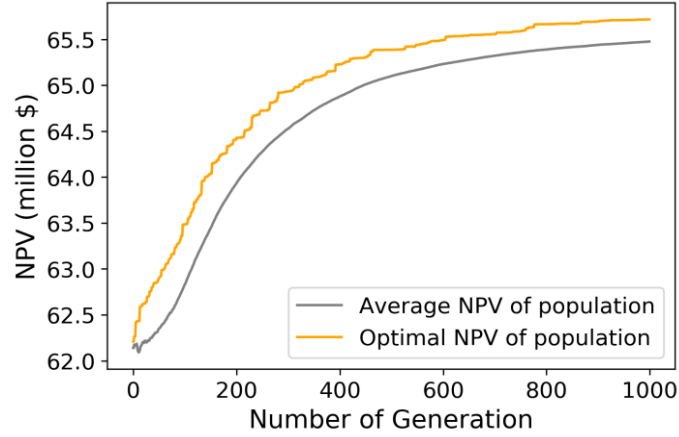

**Figure 9.** Objective function value in the optimization process.

In the optimization process with the DE algorithm, the population number of individuals $N_{ind}$ is set to be 300, and the maximum evolution generations $G_{max}$ is set to be 1,000. The well operations of producers and injectors are encoded into the individuals to update and evolve from generation to generation. The scaling factor is set to be 0.5, and the probability of crossover is 0.7. The objective function value in the optimization process is presented in **Figure 9**, and the optimization results of well control sequences for different wells are shown in **Figure 10**. To assess the effectiveness of the optimization results, we compare the NPV with the optimized well controls and 800 groups of randomly generated well controls. The comparison result is shown in **Figure 11**, in which the red line denotes the NPV obtained with optimized well controls, and the scatters denote the NPV of cases with randomly generated well controls. The blue scatters denote the unfeasible random cases, which do not satisfy the predefined production temperature constraints, and the red scatters denote the feasible random cases. A reference case has also been introduced, which follows constant well operations in the whole production process, with BHP of producers being 240 bar and rate of injectors being 1200 $m^3/day$. The NPV of the reference case is represented with a black dashed line. It can be seen that the optimized case obtains higher NPV than the reference case. Moreover, among the feasible random cases, none of them achieves a higher NPV than the optimized case, which demonstrates that the deep learning surrogate-based DE optimization can effectively identify the optimal well control operations for geothermal energy production. It is also worth noting that the optimization process only takes approximately 30 s, which is much faster than running the numerical simulators directly. Therefore, the optimization efficiency can be significantly improved by utilizing the deep learning based surrogate.

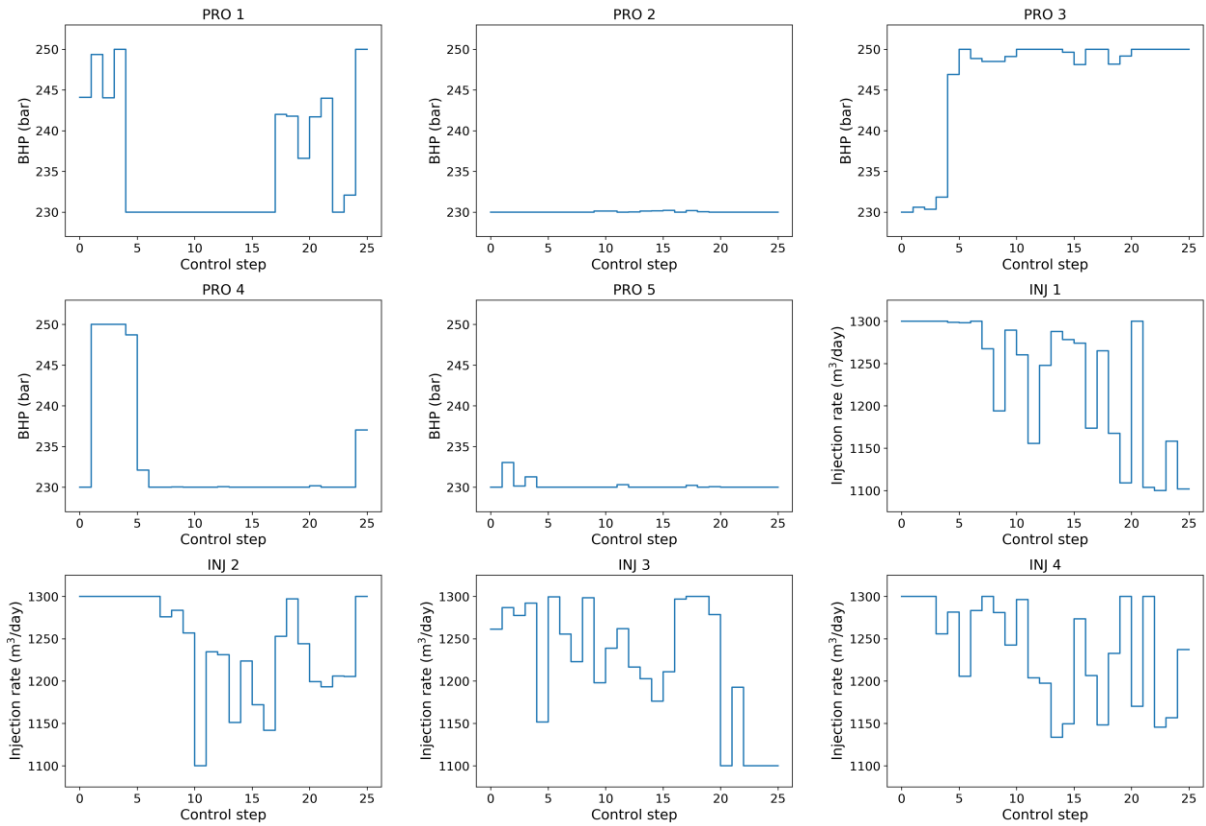

**Figure 10.** Optimization results of well control sequences for different wells.

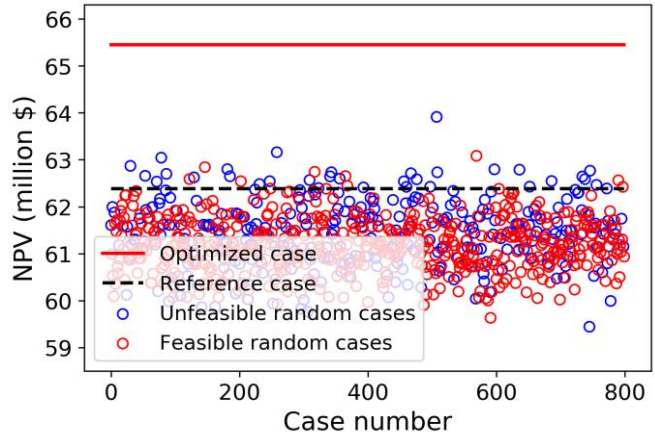

**Figure 11.** Comparison of NPV for the optimized case and randomly generated cases.

**3.2 Closed-loop optimization**

In the former case, permeability is assumed to be known, and thus geologic uncertainty is not considered in the optimization process. In this subsection, the permeability field is assumed to be unknown, and closed-loop optimization is performed to simultaneously estimate the geologic properties and optimize the well controls.

Due to the lack of information about the geologic situation, optimization would be aimless and pointless during the early stages. To avoid this, the closed-loop optimization process can be divided into two stages. In the first stage, the producers and injectors are operated to follow a predefined constant scheme, and optimization is not involved in this stage. The production data obtained from the first production stage can be employed for estimating the permeability field roughly, and the estimated geologic properties can be utilized for closed-loop optimization of the next stage. For data assimilation of the first stage, a simple deep learning surrogate can be constructed firstly to improve the inversion efficiency, in which the well controls are constant and the permeability fields are the varying model inputs. In the second stage, the closed-loop optimization workflow can be implemented to simultaneously optimize the well operations and update the geologic information. In this stage, another deep learning surrogate can be constructed, in which both the permeability fields and well controls are varying model inputs. Considering that general information about the permeability can be obtained by assimilating the production data of the first stage, the closed-loop optimization can be performed based on those estimations, and can further refine the estimations with new available production data.

Still consider the permeability field in subsection 3.1, which is assumed to be unknown in this case. The first stage is set to be the first eight time steps, in which the producers are operated to product with a constant BHP of 240 bar, and the injectors are operated to follow a constant rate of 1200 $m^3/day$. Then, the next 17 time steps constitute the second stage, in which closed-loop optimization can be performed.

### *3.2.1 Data assimilation of the first stage*

A deep learning surrogate is firstly constructed for the first stage, which can predict production data for different permeability fields, and the well controls are constant in this stage. To provide training datasets, 500 realizations of permeability fields are generated with KLE (Zhang & Lu, 2004) following the statistics shown in Eq. (9) and Eq. (10). In addition, the production data are solved with the numerical simulator for different permeability fields, which constitute the labels of the training datasets. It takes approximately 500 s to train the surrogate for the first stage. The scatter plots presented in **Figure 12** demonstrate that the trained surrogate achieves satisfactory accuracy, which can then be employed for data assimilation of the first stage.

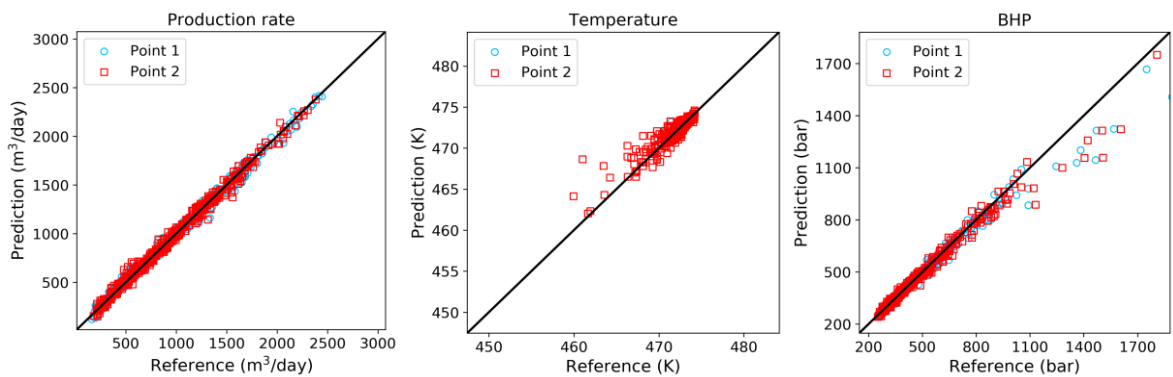

**Figure 12.** Scatter plots of production responses from surrogate predictions and references for the first stage of development.

By combining the constructed surrogate with the IES method, rough estimations of the permeability field can be obtained. In IES, 1,000 realizations are utilized to form the ensemble, i.e., $N_e$=1000, and thus 1,000 estimations of the unknown permeability field can be obtained. The inversion results of the first stage are shown in **Figure 13**, and it can be seen that the estimation is getting closer to the true reference field compared to the initial states, but a certain difference remains. Moreover, uncertainty about the permeability field has been reduced, especially at the well positions. The estimation of the permeability field can be further updated, and uncertainty can be further reduced, in the next closed-loop optimization stage.

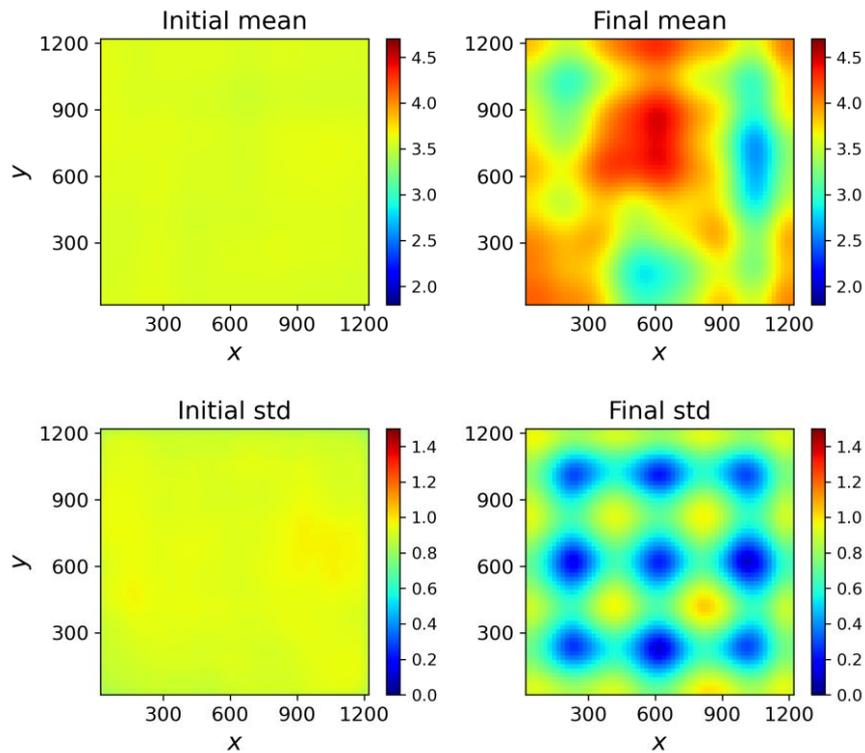

**Figure 13.** Inversion results of the first production stage.

### 3.2.2 Closed-loop optimization of the second stage

With the roughly estimated permeability fields, closed-loop optimization can be implemented for the next stage in this subsection. The deep learning surrogate is firstly constructed. To obtain the training datasets, 1,000 groups of well control sequences are randomly generated for the next stage, and the permeability fields are sampled from the group of initial estimations obtained from the first stage. The 1,000 groups of realizations are then solved with the numerical simulator to provide the corresponding production data. The deep learning surrogate can be constructed for the second production stage, which can predict production sequences (e.g., BHP, temperature, production rate, etc.) of different wells for different permeability fields. It takes approximately 1,000 s to train the surrogate for 2,000 epochs, and the prediction results for a randomly sampled realization are presented in **Figure 14**, **Figure 15**, and **Figure 16**. It can be seen that the predictions of production data can match the numerical solutions well. The scatter plots of predictions and reference values are presented in **Figure 17**, which further demonstrates the accuracy of the trained surrogate.

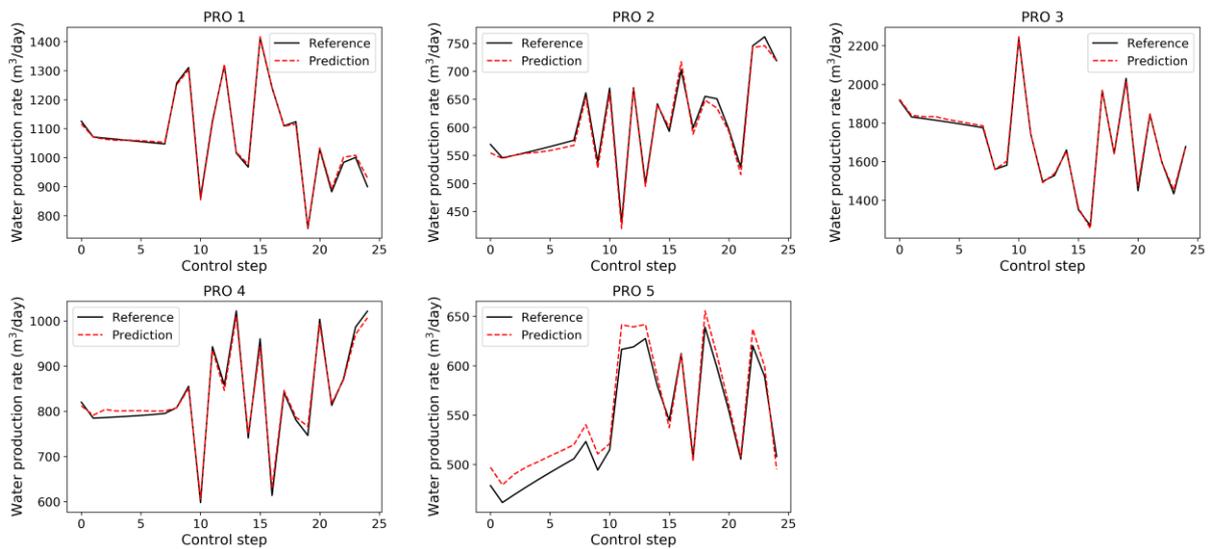

**Figure 14.** Comparisons between predictions from the deep learning surrogate and numerical solutions of water production rate.

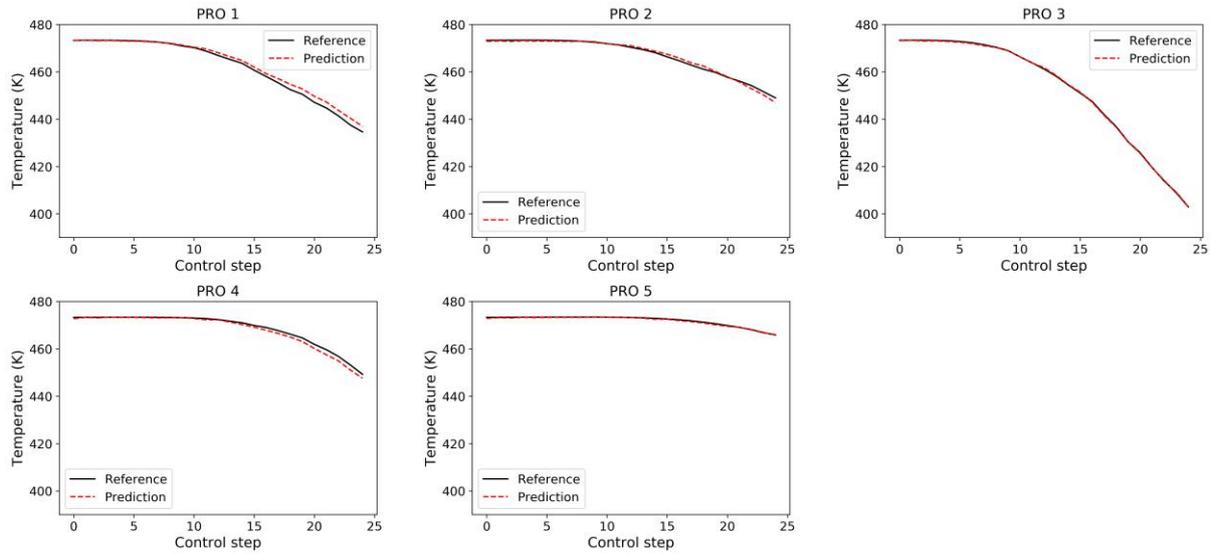

**Figure 15.** Comparisons between predictions from the deep learning surrogate and numerical solutions of production temperature.

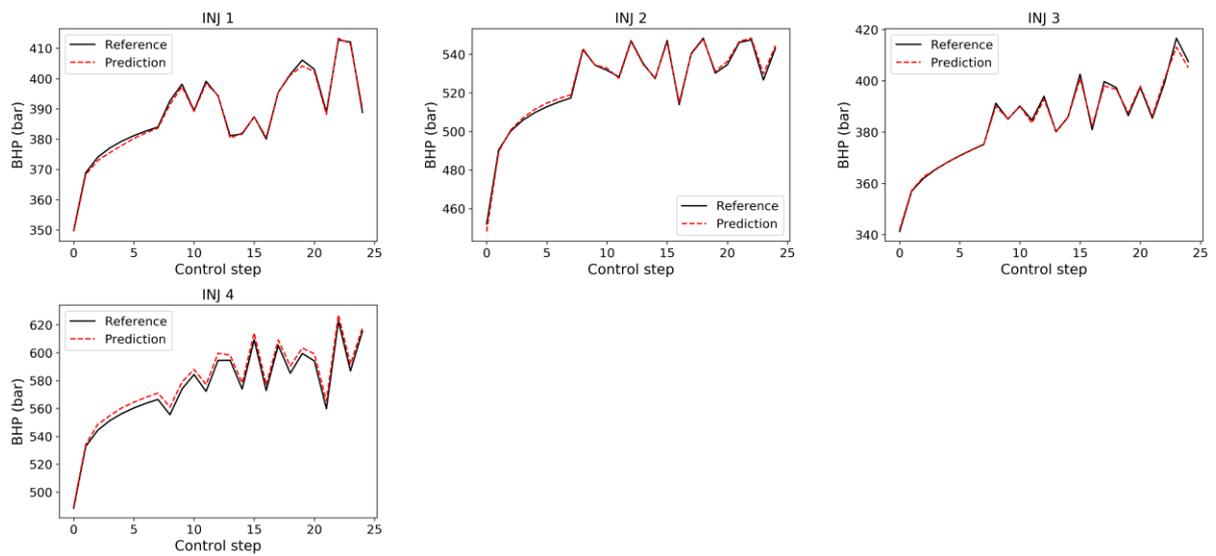

**Figure 16.** Comparisons between predictions from the deep learning surrogate and numerical solutions of BHP.

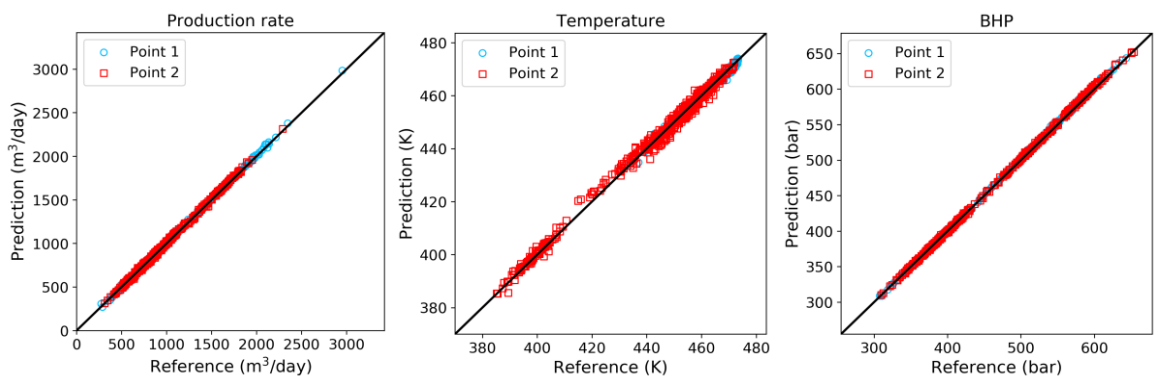

**Figure 17.** Scatter plots of production responses from surrogate predictions and references for the second

The closed-loop optimization can then be implemented with the trained surrogate model. Similar to subsection 3.1, the DE algorithm is adopted in the optimization part, and the objective function of optimization is still the NPV. The difference is that the permeability field is no longer determined, but rather associated with uncertainty. As a consequence, geologic uncertainty should be considered in the optimization process by averaging the objective function over the group of estimations:

$$\bar{O}(\mathbf{x}) = \frac{1}{N_e} \sum_{i=1}^{N_e} O(\mathbf{x},\ \mathbf{m}_i), \tag{14}$$

where $\mathbf{m}_i$ denotes the $i$th realization of model parameters, i.e., permeability field in this case. Therefore, the NPV would be calculated for each estimation of permeability field, and the group of estimations are utilized to represent geologic uncertainty, which is time-varying in the closed-loop optimization process with new data being assimilated. It is also worth noting that the optimized parameters are the well control variables for the remaining time steps of the production period, and the optimization results of the next time step would be implemented in the field. In other words, the number of control variables to be optimized is also time-varying as the production of the geothermal reservoir proceeds. After the optimal well controls are operated on the reservoir, the newly obtained production data of the new time step would be employed to update the estimations of the permeability field, and optimization of the remaining time steps can be performed based on the updated permeability field.

The closed-loop optimization results of well controls are presented in **Figure 18**. In order to investigate the effectiveness of the optimization results, 1,000 groups of well controls are randomly generated and inputted into the simulator to calculate the corresponding production data. The comparison of NPV between the optimized case and the randomly generated cases is shown in **Figure 19**. It is obvious that the optimized case can achieve higher NPV than all of the randomly generated cases, either feasible or unfeasible, which illustrates the validity of the closed-loop optimization results.

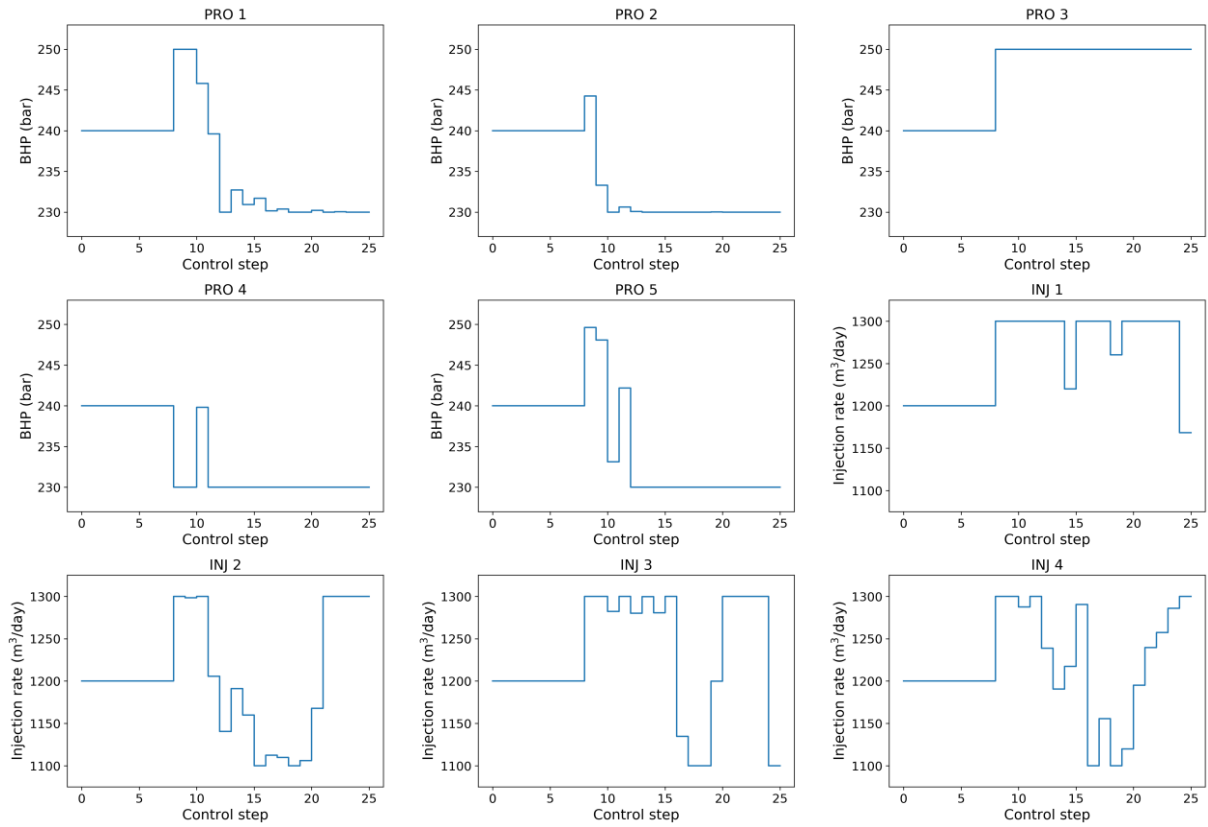

**Figure 18.** Closed-loop optimization results of well control sequences.

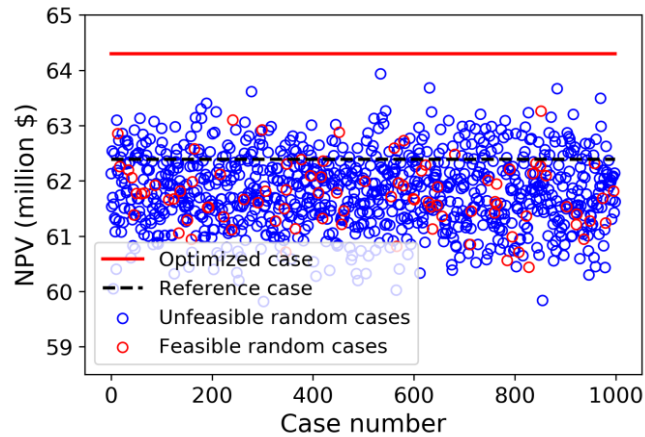

**Figure 19.** Comparison of NPV for the optimized case and randomly generated cases.

Furthermore, the characterization of the permeability field becomes increasingly clear as more production data are assimilated in the closed-loop optimization process. The estimation results of the permeability field and the corresponding uncertainty (variance distribution) in the optimization process are presented in **Figure 20**. One can see that the estimation of permeability field becomes increasingly similar to the reference field as production proceeds and more data are assimilated. The uncertainty of the estimations also decreases gradually over time. To evaluate the inversion results quantitatively, the root mean square error (RMSE) and Ensemble Spread of the estimation results are calculated with Eq. (15) and Eq. (16), respectively, which are presented in **Figure 21**.

$$\text{RMSE} = \sqrt{\frac{1}{N_{gird}} \sum_{i=1}^{N_{grid}} \left( \ln K_{ref}(i) - \ln K_{est}(i) \right)^2}, \tag{15}$$

$$\text{Ensemble Spread} = \sqrt{\frac{1}{N_{gird}} \sum_{i=1}^{N_{grid}} \sigma_{est}^2(i)}, \tag{16}$$

where $N_{grid}$ denotes the total number of grids in the discretized domain; subscripts *ref* and *est* denote the references and estimations, respectively; and $\sigma_{est}^2$ denotes the variance of the estimations in the ensemble. RMSE measures the accuracy of the estimation results, and Ensemble Spread measures the estimation uncertainty. It is obvious that both the RMSE and Ensemble Spread decrease with the progressing of the time step, which again confirms our previous results and conclusions.

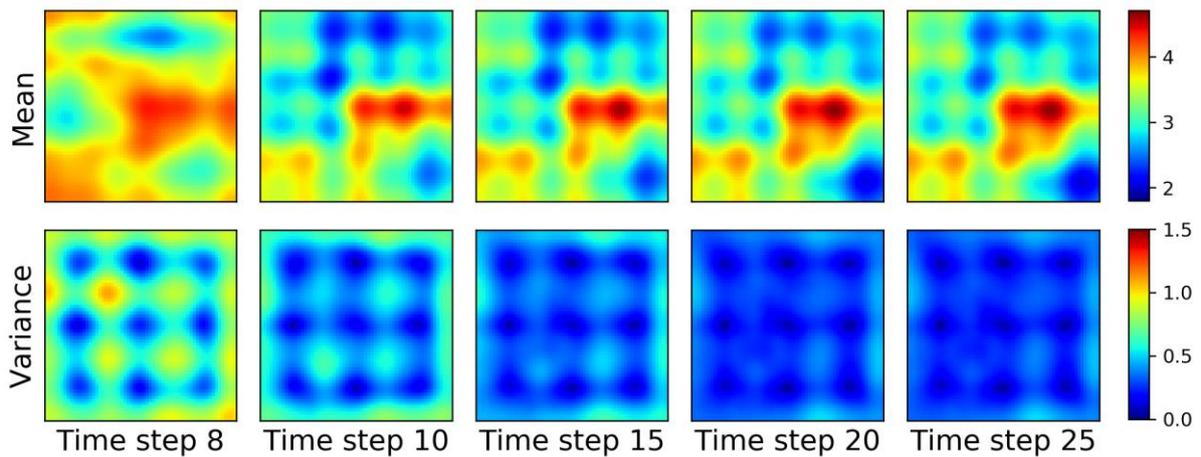

**Figure 20.** Estimation results of permeability field and the corresponding variance distribution in the production optimization process.

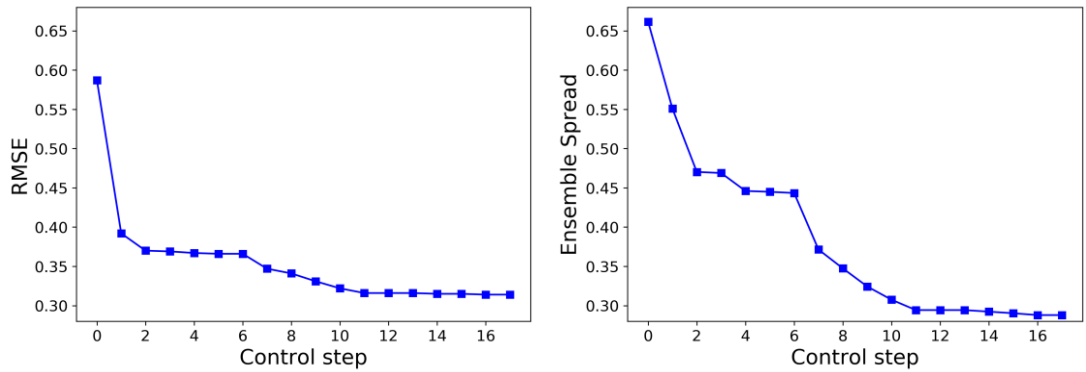

**Figure 21.** RMSE and Ensemble Spread of permeability estimation results in the optimization process.

Moreover, the rough estimations of the permeability field at the end of the first stage and the final estimations at the end of the second stage are further inputted into the numerical simulator to calculate the production results of the two stages using the optimized well controls. The production results are presented in **Figure 22**, **Figure 23**, and **Figure 24**. The solid lines denote the production results calculated using the true permeability field, i.e., the reference production data, and the error bars denote the standard deviation of the ensemble predictions. The predictions from the ensemble of roughly estimations have much higher error bars, and the median values also deviate from the references. However, the predictions from the final updated permeability fields match the references well with small error bars, and thus the uncertainty of production response predictions has also been reduced after the permeability field has been iteratively updated.

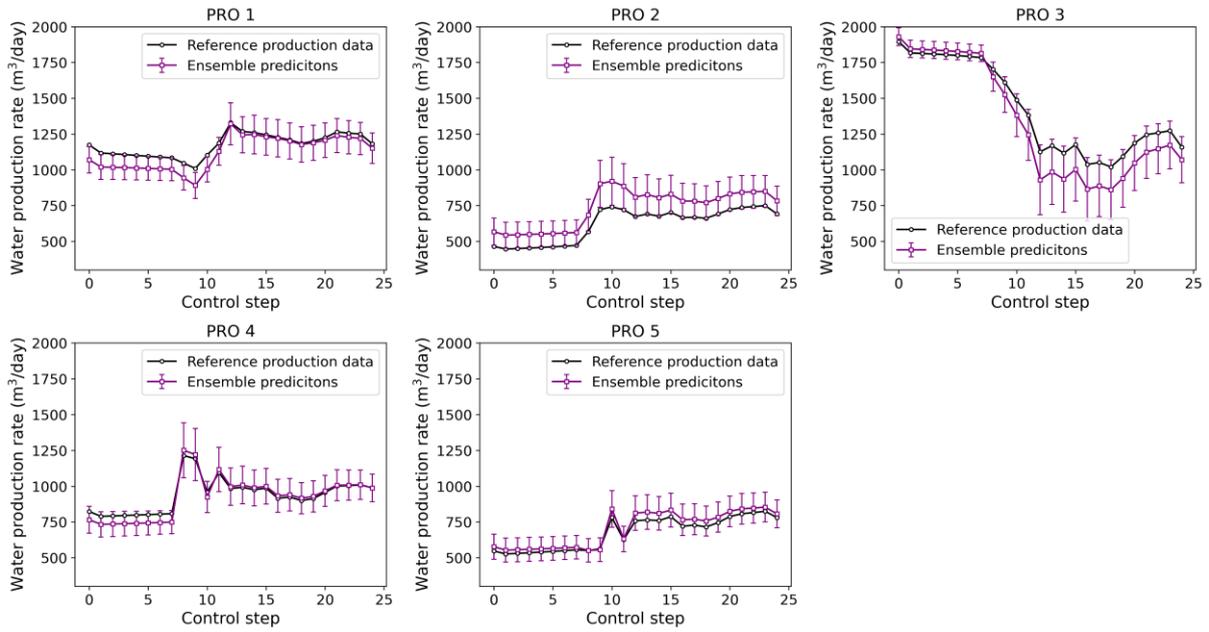

(a)

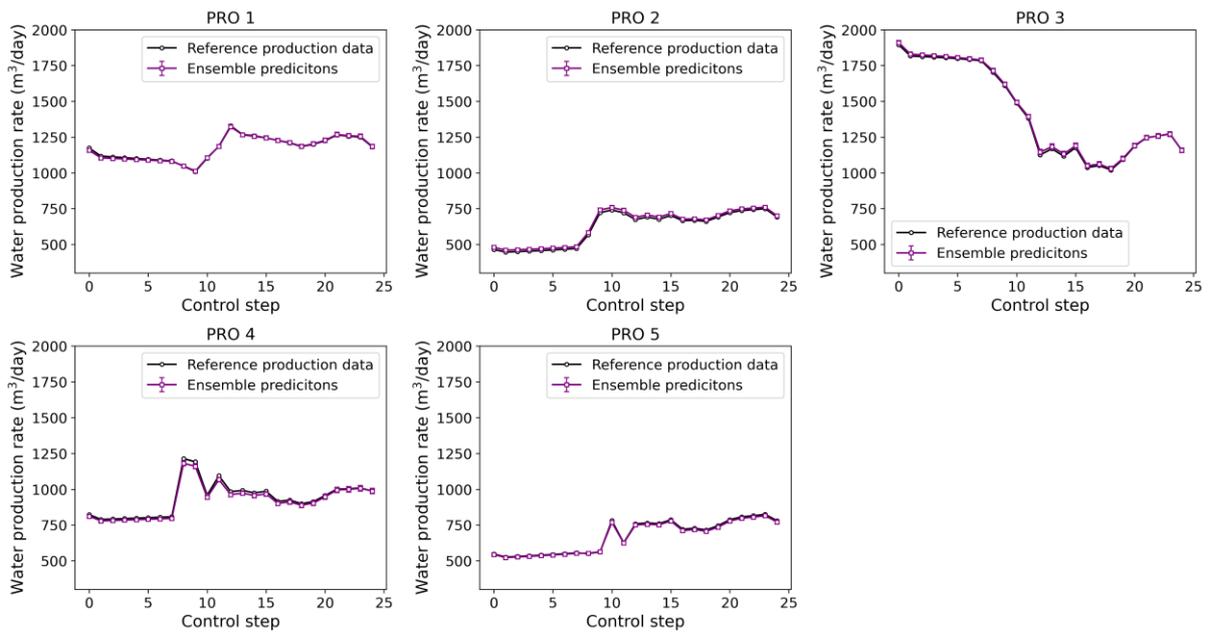

(b)

**Figure 22.** Data matching of water production rate from the permeability estimations in the ensemble (a) at the end of the first stage and (b) at the end of the second stage.

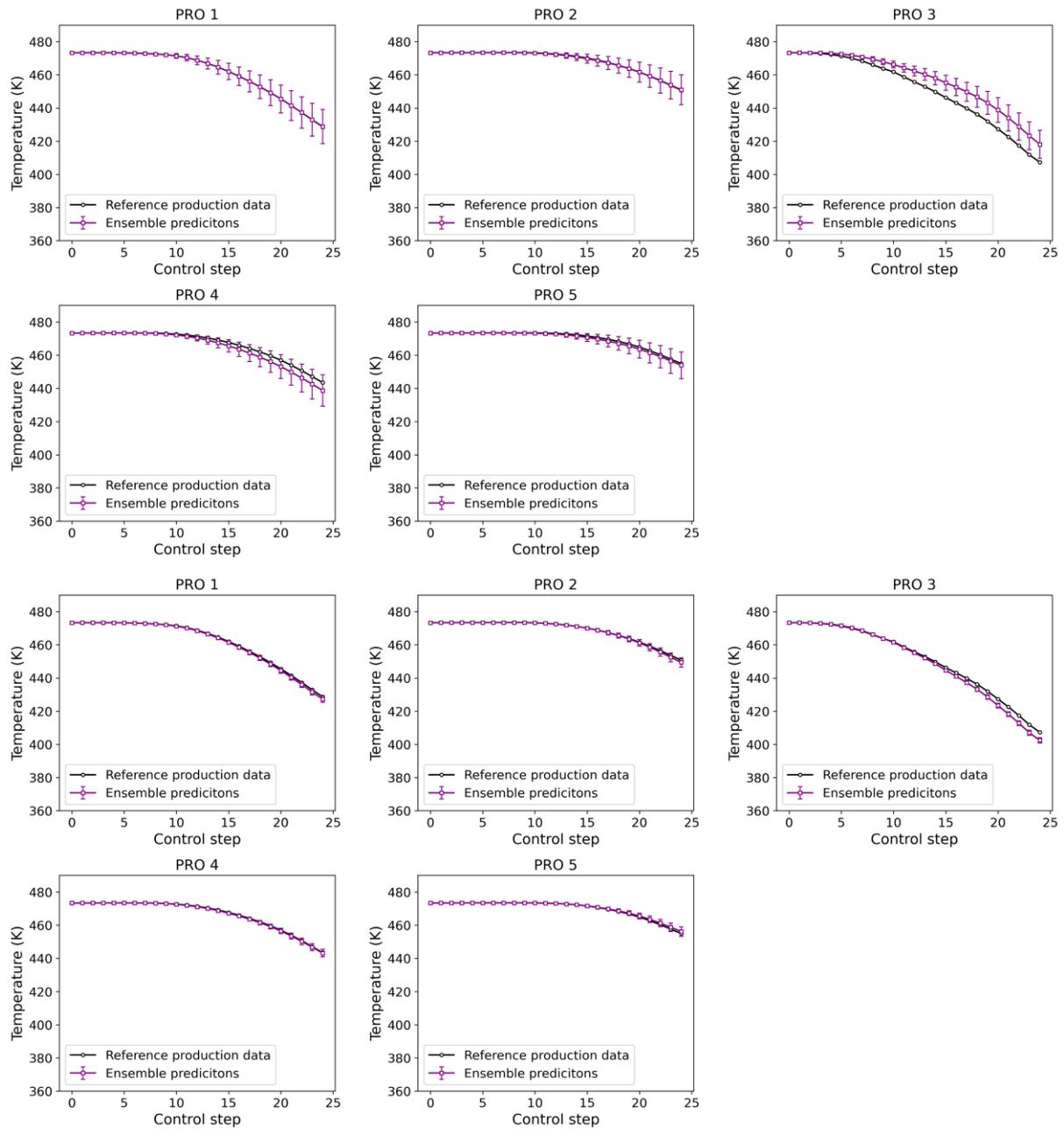

**Figure 23.** Data matching of temperature from the permeability estimations in the ensemble (a) at the end of the first stage and (b) at the end of the second stage.

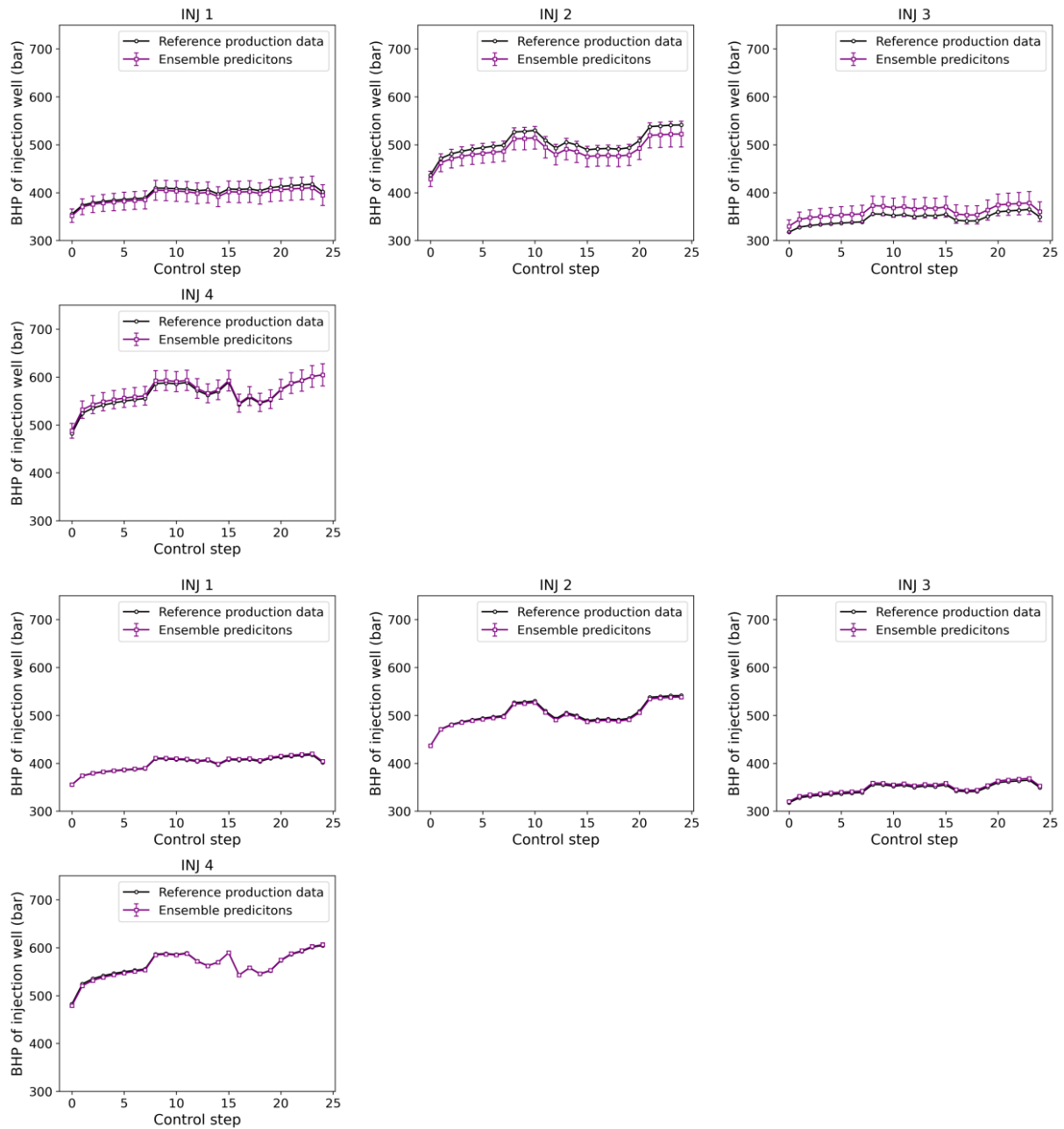

**Figure 24.** Data matching of BHP from the permeability estimations in the ensemble (a) at the end of the first stage and (b) at the end of the second stage.

## 4 Discussions and Conclusions

In this work, a deep learning based closed-loop optimization framework for geothermal reservoir development was proposed. The hybrid convolution-recurrent neural network model is constructed as a surrogate, which can extract spatial information of the parameter field through the convolution structure and approximate the sequence-to-sequence mapping via the recurrent structure. Therefore, the constructed deep learning surrogate can predict production responses for cases with different permeability fields and well control operations. The trained deep learning surrogate can then be used for production optimization. In the closed-loop optimization framework, the DE algorithm was adopted for production optimization, and the IES method was adopted for data assimilation. The optimization part can provide optimized well control operation to the geothermal reservoir, and the reservoir can provide new production data after the optimal well operations for the next control step being implemented. The newly obtained production data can be assimilated to update the geologic parameters of the reservoir, which can be utilized for production optimization of the subsequent step. Therefore, data assimilation and well control optimization can be implemented alternately in the closed-loop optimization framework.

Several geothermal reservoir development cases are designed to test the performance of the proposed production optimization framework. Firstly, the production optimization was performed for a known permeability field to validate the deep learning surrogate-DE-based well control optimization. The results show that the optimized well controls can provide much higher economic benefits than a series of randomly generated cases. The closed-loop optimization framework was then implemented on a geothermal reservoir with unknown permeability field, in which the whole production cycle was divided into two stages and two deep learning surrogates were constructed for the two stages, respectively. In the first stage, the wells in the reservoir were operated to follow a constant scheme, and the production data were utilized to roughly estimate the permeability field. Based on the initial estimations, closed-loop optimization can be performed in the second stage to simultaneously optimize the well controls and update the permeability field. The results show that the closed-loop optimization framework can provide a beneficial well control solution under some constraints. Furthermore, the estimations of permeability field become increasingly accurate, and uncertainty can be significantly reduced, as more production data are assimilated. The two-fold results demonstrate the effectiveness of the closed-loop optimization framework for geothermal reservoir production. In addition, through the assistance of the deep learning surrogate, closed-loop optimization can be implemented with high efficiency.

**Acknowledgements**

The first author gratefully acknowledges support from the China Scholarship Council scholarship (No. 202106010163). This work is partially funded by the Shenzhen Key Laboratory of Natural Gas Hydrates (Grant No. ZDSYS20200421111201738) and the SUSTech - Qingdao New Energy Technology Research Institute.